\documentclass{article}

\usepackage{arxiv}

\usepackage[utf8]{inputenc} 
\usepackage[T1]{fontenc}    
\usepackage{hyperref}       
\usepackage{url}            
\usepackage{booktabs}       
\usepackage{amsfonts}       
\usepackage{nicefrac}       
\usepackage{microtype}      
\usepackage{lipsum}

\usepackage{graphicx}
\usepackage[cmex10]{amsmath}
\usepackage{amssymb}
\usepackage{algorithmic}
\usepackage{array}
\usepackage{setspace}
\usepackage[caption=false,font=footnotesize]{subfig}
\usepackage{float}
\usepackage{url}
\usepackage[noadjust]{cite}
\usepackage[toc,page]{appendix}
\newcommand{\pindent}{0.4cm}

\title{Deep Learning Regression of\\VLSI Plasma Etch Metrology}

\author{
Jack Kenney \\
College of Computer and Information Sciences\\
University of Massachusetts, Amherst\\
\texttt{jnkenney@umass.edu} \\
\And
John Valcore\\
Lam Research Corporation\\
Fremont, CA 94538 \\
\texttt{john.valcore@lamresearch.com} \\
\AND
Scott Riggs \\
Lam Research Corporation \\
Fremont, CA 94538 \\
\texttt{scott.riggs@lamresearch.com } \\
\And
Edward Rietman \\
College of Computer and Information Sciences\\
University of Massachusetts, Amherst\\
\texttt{erietman@umass.edu} \\
}
\date{May 2019}
\begin{document}
\maketitle
\begin{abstract}
In computer chip manufacturing, the study of etch patterns on silicon wafers, or metrology, occurs on the nano-scale and is therefore subject to large variation from small, yet significant, perturbations in the manufacturing environment. An enormous amount of information can be gathered from a single etch process, a sequence of actions taken to produce an etched wafer from a blank piece of silicon. Each final wafer, however, is costly to take measurements from, which limits the number of examples available to train a predictive model. Part of the significance of this work is the success we saw from the models despite the limited number of examples. In order to accommodate the high dimensional process signatures, we isolated important sensor variables and applied domain-specific summarization on the data using multiple feature engineering techniques. We used a neural network architecture consisting of the summarized inputs, a single hidden layer of 4032 units, and an output layer of one unit. Two different models were learned, corresponding to the metrology measurements in the dataset, Recess and Remaining Mask. The outputs are related abstractly and do not form a two dimensional space, thus two separate models were learned. Our results approach the error tolerance of the microscopic imaging system. The model can make predictions for a class of etch recipes that include the correct number of etch steps and plasma reactors with the appropriate sensors, which are chambers containing an ionized gas that determine the manufacture environment. Notably, this method is not restricted to some maximum process length due to the summarization techniques used. This allows the method to be adapted to new processes that satisfy the aforementioned requirements. In order to automate semiconductor manufacturing, models like these will be needed throughout the process to evaluate production quality.
\end{abstract}

\keywords{Machine learning \and VLSI \and Plasma Etch \and Metrology \and Nanoscale}

\section{Introduction}

The manufacture of computer chips uses a plasma reactor that contains an ionized gas to conduct a series of chemical operations on the surface of a silicon wafer. Prior to being placed in the reactor, a mask is applied to the wafer in the shape of a desired circuitry pattern. Inside the reactor, various substances are deposited on the wafer and activated away from areas that do not have the mask applied, producing trenches in the surface of the wafer. The sequence of depositions and activations is defined in a recipe, the execution of which will henceforth be referred to as a process. The success of a process is determined by measuring different aspects of the etched patterns, which is called metrology. To ensure that processes have been successful without performing costly tests, such as taking a cross-sectional image of a finished chip, we would like to be able to predict the metrology measurements using automatic tools.

The dataset was comprised of inputs of environmental sensor information from a plasma reactor with a wafer inside during a Very Large Scale Integration (VLSI) process. The outputs for the process are the measurement values derived via metrology on cross-sectional images of a chip taken from the corresponding etched wafer. Across the dataset, each chip was taken from the same location on its wafer to reduce variability. The goal of this project was to apply machine learning (ML) regression models to map the sensor data as input and the metrology as output.

To address this regression problem, we first modified the data using a feature engineering pipeline. Data expansion techniques for simulating a larger dataset based on information lost in the data compression were considered, but were not found to provide significant improvement. We then trained a neural network with the technique of grouped leave-one-out cross-validation (LOOCV) on the engineered dataset to determine the hyperparameters of the models that are best suited to the data and tested the final hyperparameters on a held-out test set to estimate the generalization error. This method is used to avoid making a multiple comparison procedure and overestimating the efficacy of the model.

Overall, our goal was to make accurate predictions of process outcomes with the provided data, using various statistical and ML methods. A critical feature of this work was that, given the small dataset size and limitations, the results and methodology were as statistically sound as possible, according to ML standards at the time of writing.

\section{Significance}

Plasma etching is ubiquitous in semiconductor manufacturing because the plasma, an ionized etching gas, can be directed into the material substrate to be etched by the chemically active ions. Similarly, plasma can used for deposition. The two processes can be combined for complex submicron and nanoscale structure fabrication, often with scores of processing steps. Every etch process has specific chemistry and instrument setpoints including radio frequency (RF) power(s) for creating and maintaining the plasma, direct current bias on the wafer surface, mass flow control of various gasses, pressures, and many other factors. 

A statistical design of experiments on a set of wafers is often conducted to identify the required conditions and settings for an etch process, followed by fine-tuning the control parameters using another set of test wafers before the process can be used in manufacturing. Plasma etching is a very complicated and costly process that can be optimized using various machine learning methods, one part of which is the goal of this project.

A skilled engineer will be able to guess the approximate end results, that is, the etched structures and features on the wafer, from knowing the recipe. A first step in applying machine learning to streamline research and design for manufacturing processes is to predict the final etch results as would be observed on the wafer surface by metrology tools, from environmental sensor data. To this end, there have been many papers in the literature describing both the end result of an etch and virtual metrology modeling for plasma etch operations. Early examples include \cite{rietman1993use, rietman1996neural, kim1994optimal, kim1997real, kim2004prediction, kim2007prediction}. More recently are \cite{kim2009modeling, zeng2009virtual, lynn2012global}.

Given the vast array of chemical processes, considering both etch stack and gases used in deposition and etching, it is not reasonable to expect any single neural network, or other instance of a machine learning algorithm to generalize beyond the chemistry it was trained on. Nonetheless, as cited in the literature above, plasma etch modeling and/or metrology prediction can be used for real-time control of manufacturing processes, for failure prediction, and automated process analysis.

Predicting end-of-process results from time-series data is a difficult task in modern machine learning, especially when data is collected at various fidelity. The problem is particularly difficult when processes have varying lengths, such as is the case with the raw data with which this project began. Making appropriate feature engineering choices to accurately summarize and consolidate sensor data takes domain knowledge, planning, and careful execution. In the commercial field of VLSI silicon processing, the cost of performing experiments is high and interferes with manufacturing efficiency. Thus, being able to predict results from \textit{in situ} process sensors is highly valuable.

Due to the cost of gathering the samples used in this project, the number of examples for the model to learn from is low, which makes the problem even more difficult, pushing the boundaries of modern machine learning and engineering. Being able to predict the critical dimension (CD) measurements of the etched silicon with a low enough error, using only process sensors, we take a step forward in automating the silicon etch process, saving time, money, and resources in the industry that ubiquitously underlies modern computing technologies.

The research question we are addressing pushes the limits of machine learning under extreme conditions. In this case, we have very few, 14 in total, examples from the distribution of etched wafers in the recipe class we are addressing. In the raw data, there are millions of features associated with each wafer, and up to eight metrology measurements to regress. With so little data, the process by which we feature engineer and examine model efficacy is crucial to the project's success. Notably, the proper execution of this task will result not in an ideal system model but instead serve as a proof-of-concept in plasma etching of complex micro-structures with multilevel-materials. In the particular process we studied, there were scores of activation and deposition steps repeated, the number of which will only increase as the industry progresses. To accommodate this, our feature engineering techniques are designed to summarize across the \textit{type} of step, e.g. deposition or activation, compressing the number of times that type was repeated into a fixed number of parameters corresponding to a polynomial fit. This method increases the number of processes our methodology could apply to, including those of differing lengths while supporting extremely large step counts.

\section{Methodology}

The steps are initially described by time-series data for various sensor types. The data can be broken down into known segments, called cycles, each of which corresponds to a particular action. For example, the fifth cycle of depositions is the fifth time that a deposition has occurred in the process. To compress the data from its raw form, we decomposed it into fitted parameters over the time within each cycle and then fitted those parameters according to its corresponding type of cycle, or step, such that they were of a reasonable dimension to be used as input for machine learning models. Finally, dataset of inputs-to-outputs was constructed, aligning the input sensor data appropriately with the measurements taken from the corresponding wafer images as output. The images were taken using a Scanning Electron Microscope and then measured by an engineer. The dataset was then used in a cross-validation procedure to identify the best model for the problem and estimate generalization error.

\subsection{Process Overview}

\begin{figure}[H]
    \centering
    \includegraphics[width=0.75\textwidth]{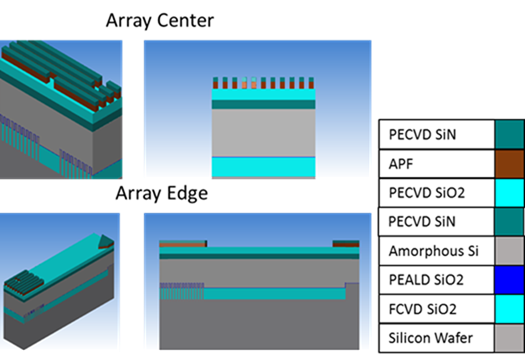}
    \caption{The Etch Stack: An example wafer construction with a complex etch stack used in etch processes.}
    \label{fig:etch_stack}
\end{figure}

\hspace{\pindent} The dataset was derived from a plasma etch process outlined by a specific wafer structure, procedure, and sequence of gaseous chemistry. The wafer material stack is displayed in Figure \ref{fig:etch_stack}. We used a surrogate structure as a test vehicle for a non-self limiting atomic layer epitaxy (ALE) application. The ALE was accomplished by depositing a passivation layer onto the wafer surface using chemistry $A$. The purpose of $A$ was to generate molecules that will attach to the film(s) of interest. Chemistry $A$ with energy $E_A$ was followed by activating the now-passivated surface with chemistry $B$ at energy $E_B$, where chemistry $B$'s primary purpose was to create ions to facilitate reactive ion etch (RIE). The surrogate structure we used was a line and space grating mask. The film of interest we were etching was an oxide, and film we did not want to etch will be referred to as non-etch-material (NEM). The $A$ was a fluorine compound, and the chemistry of $B$ contained $Ar$. The chemical depositions and activations were divided into a complex sequence of several repeated steps which allowed for deep nanostructure fabrication. Most of the steps were deposition of materials and activation etching steps, but there were others for chamber environmental stabilization and initial striking of the plasma, which were agnostic to process on-wafer.

After the plasma etch process was completed, a die, a small area similar to an integrated circuit, was cut from the same place on each wafer, notably in the center to reduce edge effects. The die was then cut cross-sectionally in the same location for image metrological analysis, and important critical features were measured and recorded by a domain expert process engineer. These measurements served as output for the regression machine.

\subsection{Dataset}

The dataset consists of fourteen whole wafer runs. Each wafer has a collection of sensor measurements and a set of corresponding metrology measurements. The layered chemical composition of the wafers corresponds to Figure \ref{fig:etch_stack}. Wafers R17, R19, R20, R21, R22, R23, R24, and R25 nominally came from the same chamber condition, meaning they were processed consecutively with minimal RF hours between wafers. Because they are similar, we will name them collectively Group 1. Running the wafers in close sequence is important because the chamber undergoes many changes during the etch process and the conditions change as etching occurs. These wafers being run in sequence means that the difference in their signals is more likely due to the variation that is occurring in the etch itself rather than the changes in the chamber that have occurred during unrelated etches. These represent the bulk of the data and are considered relatively normal.

Wafers R28, R29, R30, R31, on the other hand, were run after the chamber had a full wet clean where various chemicals were run through the interior of the reactor to remove buildup from previous etching processes. Consumable parts were also replaced. We will refer to that set of wafer collectively as Group 2.

Wafers R32, R33, and R34, were run after those changes and were from a new lot of wafers from the supplier. The new box of wafers from the supplier had slightly smaller stack structure. As the initial wafer geometry changed, the final output geometry is expected to be different as well, so we will refer to those as Group 3. 

Mapping the raw recipe setpoints to the output measurements alone, without counting for pre-etch geometry, could possibly introduce error into the model, but due to the limited dataset and the variety that the final model would be expected to generalize to, all wafers were used. The Group 3 wafers were also run after Group 2, meaning the chamber condition was not the same as for Group 1.

By taking sequential samples before and after a wet clean, we aimed to build a predictive model more robust to changes in chamber condition over time as well as with varied parts. As such, we conclude that though the data was not truly independently and identically sampled, the variation of the samples was considered to span much of the distribution that future wafers from the same process type, those to which the model will be expected to generalize, are going to be drawn from. 

To test this hypothesis, we selected wafers R22 from Group 1 and R34 from Group 3 for testing, as they represent very different sections of the dataset: the former was early in the process batch, and the latter came after the wet clean and from a different lot of wafers. The results from testing on these wafers is discussed in Section \ref{sec:results}. In this way, the experiment is reasonably directed toward application with models that carry implicit assumptions of independence among training and testing samples.

The recipe produces data that occurs in multiple steps, the most important of which are the etch steps. We focus on those steps for the problem, because the deposition and activation steps are the most relevant to the etching process and the other steps were not considered important. The sensor data is compressed by selecting a subset of the sensor variables, calculating linear fit coefficients across each cycle, and then fitting a third-order polynomial to those calculated coefficients across the \textit{cycles} of the each distinct step type. We selected only deposition steps and activation steps, due to variety in the recipes in the data set, and ignored non-etch steps.

\subsection{Data Preprocessing}

To take the raw data files from the reactor and produce meaningful samples for a neural network to consume, the order of the data needs to be dramatically reduced. There are a set of wafers $W=\{w_1\cdots w_r\}$, where $r=14$. Originally, the data files are on the order of 1,000 columns and 50,000 rows. The columns were represented by a set $V=\{v_1\cdots v_c\}$, where $c$ was the number of columns, we selected some $c'$ variables to use in creating an example.
To further reduce the quantity of features per example, we defined a set $S=\{s_1\cdots s_p\}$, where $p$ was the number of critical steps in the recipe. For this etching process, the engineers determined that focusing only the activation and deposition steps is sufficient to learn meaningful relationships. Each step was repeated $q$ times, which defines a set of cycles $C=\{c_{11}\cdots c_{pq}\}$that we used to compress the rows of the example. To do this, we considered each cycle $c_{ij}$ separately. 
$\forall i: \forall j:$ we trim the starting values to a critical window and collect $A = \{m, b, f\}$, which are the slope and intercept of the linear fit and the approximated asymptote of the cycle, respectively, these will be referred to collectively as the ``intracycle augmented linear fit coefficients.'' Henceforth, we will indicate the $A_k$th fit coefficient for cycle $c_{ij}$ using a superscript, e.g. $c_{ij}^k$ as seen in Equation \ref{eq:L}, and in in the presence of only a step $s_i$ as some coefficient $x_{ik}$ as seen in Equation \ref{eq:P}. 

\begin{equation}\label{eq:L}
    \forall i\in[0,p): \forall j\in[0,q):\quad L(c_{ij}) = c_{ij}^b+c_{ij}^m (t+o)\\ c_{ij}^f = \frac{1}{N}\sum_{i=(1-l)N}^N c_{ij}(t+o) 
\end{equation} 

where $t$ is time and $o$ is the start of the cycle including trimming from the beginning of the time series, $N$ is the length of the cycle, and $l$ is the percent of the cycle to consider representative of the final magnitude of the cycle, which was in most cases 10\%.

Considering one distinct augmented fit coefficient $A_k$ at a time, we fit the cycles for a distinct step, gathering coefficients referred to in Figure \ref{fig:effective_model} as the ``stepwise intercycle polynomial fit,'' as follows:

\begin{equation}\label{eq:P}
\forall s_i\in S: \forall j\in[0,q): P(c_{ij}^k) =  a_{im} + b_{ik}j + c_{ik}j^2 + d_{ik}j^3
\end{equation}

Finally, with these $c'\times |S|\times |A| \times |P|$ coefficients and using our domain knowledge of the etch process our data is sufficiently summarized and is ready for input into the neural network.

Unmentioned above in Equation \ref{eq:P}, we additionally added a weighting scheme to the first few cycles $c_ij, \forall j\in [0,2]$ of the data in a 10\%, 20\%, and 70\% ratio because those cycles were determined in preliminary analysis to be unrepresentative of, and less influential than, rest of the cycles later during the process. It combines the three data into a single data point for each function of the cycles coefficients. This was due to the reactor being still in its initial phases of startup during those cycles.

For output metrics, cross-sectional images were taken of each wafer and metrology measurements calculated by Lam engineers on each of the five to six images of the die taken. Each image was measured on four metrics. Critical Dimension Trench is the width of the part of the surface that is being etched. Critical Dimension Trench Mask is the width of the part of the surface that has a mask. Recess the depth of the trench that was etched. And lastly, Remaining Mask is the amount of residual mask that is still present on the surface of the wafers. The two measurements we focused on in the prediction problem were Recess and Remaining Mask, as the other two are less deterministically related to the sensor data collected and more related to the geometry of the feature mask applied to the surface of each wafer.

\subsection{Neural Network}

\begin{figure}[h!] 
    \centering
    \includegraphics[width=\textwidth]{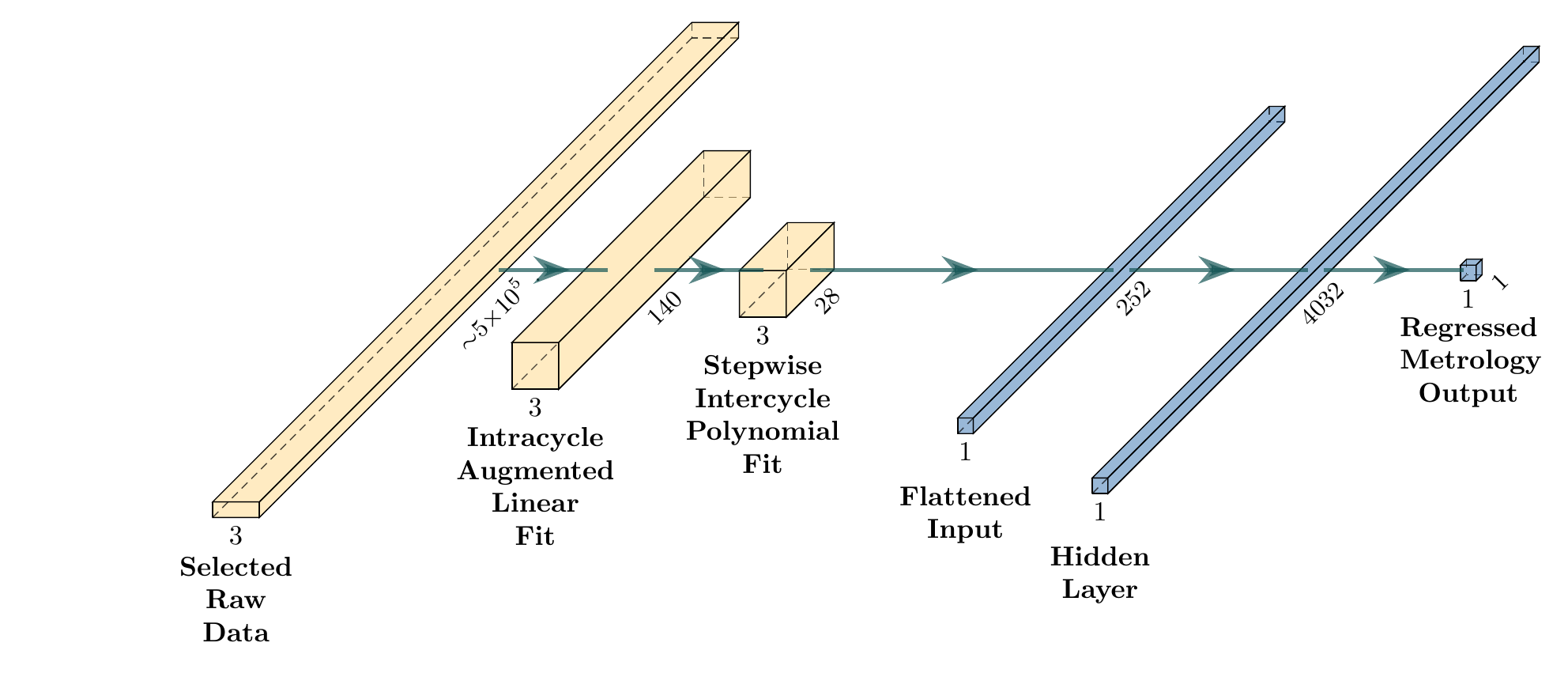}
    \caption{The Effective Model: In this diagram, we see the effective regression model. The transitions between the first three layers are static transformations from the raw reactor data. First, the input data is converted into two linear fit coefficients over each process cycle. We also augment these coefficients by including the final magnitude of each, which carries information about the asymptote of the cycle. See Equation \ref{eq:L}. This generalizes the model to incorporate time-series data of any length, when it is divided into cycles, a common procedure in etch processes. The second set of transformations apply a third-degree polynomial to the augmented linear coefficients over each distinct type of cycle, or step, see Equation \ref{eq:P}. The fourth layer is the unraveled vector version of the third layer, and is the input to the neural network model. The last three layers are part of the model with learned weights. There are 4032 hidden units in the fifth layer, which is a learned linear combination of the polynomial coefficients. After the activation is applied, the hidden units are again linearly combined into the final numeric output, using learned weights. This diagram was created using a modified version of \cite{haris_iqbal_2018_2526396}.}
    \label{fig:effective_model}
\end{figure}

To determine the appropriate neural network architecture for the features we engineered, we performed cross-validated search over various hyperparameters. Specifically, we used LOOCV on the wafers, meaning that it was twelve-fold cross-validation over the training set. The loss function used for evaluating the efficacy of a model was negative mean squared error (NMSE), which we use because we are regressing real valued numbers and is provided standard in the libraries used for the project. In the next section, we discuss the cross-validation procedure, followed by the hyperparameter search for the neural network. In Figure \ref{fig:effective_model}, the ``effective model'' is described, which is the full data pipeline represented as a deep neural network where the first three layers are considered frozen, static transformations of the input data, and the last three layers correspond to learned weights, as determined by the details in the next two sections.

\subsection{Leave-One-Out Cross-Validation Procedure}

In order to ensure that the experiment was statistically sound, the cross-validation procedure was carefully constructed to preserve the validation sets as strictly excluded from their corresponding training sets. When the dies were cross-sectioned, five to six images were taken and output measurements were taken for each of the images. This means that for a given input vector, there are five to six corresponding output examples, which are related to one another. To avoid a situation of testing on the training set, each wafer was given a label, and all of the five to six examples corresponding to that wafer were used as either training or validation data, but not both. Only one wafer, including all of its sub-examples, was used for validation at a time, thus LOOCV was used, despite having multiple samples per validation set. This strict partition of data ensured the statistical viability of the cross-validation procedure for this experiment.

When batch normalization \cite{goodfellow2016deep} was used to normalize the training and validation sets, specifically using the \textit{training set} as the baseline, not the validation set.
\begin{equation}\label{eq:norm}
    \forall v_i \in V: (v_i - \mu_T) / \sigma_T
\end{equation}
In Equation \ref{eq:norm} above, T is the set of examples in the training set and V is the set of examples in the validation set. This is statistically sound, because there is no implicit information about the mean and variance of the validation data provided in the normalization, and as mentioned before, the validation wafers are considered to have been drawn from the same distribution as the training set. Thus, using the mean and variance of the training set as an estimate of the mean and variance of the distribution is sound, within some level of confidence. 

When data expansion techniques \cite{goodfellow2016deep, geron2017hands} were used, the training data consisted of the expanded data and the validation data consisted of the non-expanded original data, because testing on expanded data is not what the model will be expected to do in the future. The expanded data were generated by sampling from a Gaussian distribution that used the covariance matrices of the fit parameters in Equations \ref{eq:L} and \ref{eq:P}, where the mean and variance were each calculated using a hyperparameter that specified the sampling percentage of the standard deviation.

Unfortunately, neither the batch normalization nor the data expansion techniques proved fruitful in improving model generalization error. The full table of cross-validation training errors, validation errors, and their corresponding measurement errors from the supervised learning process is available in Appendix \ref{appendix:errors}.

\subsection{Hyperparameter Search}

In the search for the model that would generalize best on future examples, we considered the number of layers involved, number of hidden units in those layers, activation function, amount of L2 regularization to apply to the weights \cite{geron2017hands}, learning rate, decay of the learning rate, loss function, number of epochs, and early stopping \cite{goodfellow2016deep} sensitivity in terms of minimum threshold and patience. The full number of hyperparameter combinations attempted was not exhaustive, as the process would be too resource intense. Due to the limited size of the dataset, however, we were able to be more exhaustive in our search of the hyperparameter space than is typical for deep learning projects. Usually deep learning projects utilize massive amounts of data to learn the weights of many layers and a larger feature input space, but due to the frozen transformations of the input data we are only learning a relatively small single hidden layer automatically and thus many hyperparameter settings were able to be tested. The full list of parameters tested can be found in Appendix \ref{appendix:hypers}.

The architecture that best approximates the solution to this regression problem had an input layer of size 252, one hidden layer of size 4032, and a single number regressed as output. The hyperbolic tangent activation function was used to add non-linear capabilities to the model. The L2 regularization constant was 100, which helped prevent the model from overfitting to the training data. The learning algorithm used was stochastic gradient descent with a learning rate of 1e-5 and a simulated annealing decay value of 1e-8, which finds a set of weights that minimizes the prediction error. It used a batch size of 32 to train the model over 100 epochs. On certain examples, early stopping, which helps prevent overfitting by saving and using a model trained up to an epoch before the 100th, if it is better than the 100th. The minimum validation error delta for early stopping was set to 0 with a patience of 10 epochs. This allows for noise not to cause early stopping to terminate training too early. To see the learning curves for each fold in the final two versions of the neural network, see Appendix \ref{appendix:curves}.

\section{Results}\label{sec:results}

When considering the relative success of this project, it is important to consider the ability of the model to generalize to \textit{new} data from the same process and thus demonstrate its ability to learn. Our constrained version of LOOCV allowed us to perform hyperparameter search to select hyperparameters that will generalize well to unseen instances of wafers from this production process. By performing LOOCV, we limit the bias that is introduced when hyperparameters are selected from validating on only a small subset of the data. Instead, the method uses every part of the training dataset for validation. This limits the amount of influence that a single validation wafer would have on the chosen hyperparameters. To estimate the error rate of the model on new data, we held out two testing wafers thought to be from very different sections of the distribution. For each of the output variables we focused on, we computed testing error and the corresponding measurement error on each wafer.

Mean Squared Error obscures the units of the error because it is squaring the error and taking the mean over many samples. Technically the units are $nm^2$, but this does not translate to a tangible quantity. The measurement error metric is more meaningful because it is the average error \textit{in nanometers} among predictions of output variables on unseen data. Using the appropriate hyperparameters in Table 1, we are able to accurately regress the Remaining Mask and Recess output measurements \textit{approaching the sensitivity of the imaging machine}.

\begin{table}[H]
\centering
\begin{tabular}{|l|c|c|c|}\hline
\textbf{Variable} & \textbf{Wafer} & \textbf{Testing Error (MSE)} & \textbf{Measurement Error (nm)} \\\hline
Remaining mask  & R22 & 9.11268 & 2.73841 \\\hline
Remaining mask  & R34 & 0.31283 & 0.44993 \\\hline
Recess  & R22 & 191.34170 & 13.73289 \\\hline
Recess  & R34 & 194.64074 & 13.87892 \\\hline
\end{tabular}
\caption{Estimated generalization errors for two separate models, each regressing a different output variables on the two wafers kept in the held-out test set that originate from the same process the models were trained on.}
\label{table:generalization}
\end{table}

As we can see in the above table, there is a much larger error in predicting the value of Recess. This is thought to be predominantly caused by two factors: first, the scale of the Recess variable is much larger than the scale of the Remaining mask variable, and second, as seen in Appendix \ref{appendix:curves}, the hyperparameters resulted in training curves that are much more ideal for the Remaining mask model than for the Recess model. 

\section{Conclusion}

The prediction of process results in general is a difficult task, particularly in the context of plasma etch integrated circuit fabrication, in which the 3D plasma behavior is too complex for current physics modeling techniques. Running experiments to empirically understand the plasma conditions is too costly and time-consuming. And, with such a small dataset as was provided for this project, the limits of the efficacy of ML were being tested. 

Despite the constraints on the project, we feel the results are promising and that, especially with a larger dataset, a full ML prediction pipeline, which ingests sensor data from the reactors, makes predictions, and prescribes recipe adjustments, could be implemented first in research settings and then in production to aid skilled engineers in refining recipes for specific wafer structure, mask design, and reactor settings.

In general, the model is robust to variable-length time series data without requiring complicated neural network architectures. From a data collection perspective, the only requirements for this model to be trained or predict on a new instance is that the data generating process contains the three sensor variables we isolated and contains multiple cycles of the seven steps identified as influential in predicting the outputs. The robustness of this model is particularly convincing due to the sparseness and variety of the sample from which the model was trained, paired with its relative generalization error in predicting the measurements of the post-production wafer seen in Table \ref{table:generalization}. 

These factors considered, the project can be concluded to have succeeded in its goal of utilizing a small dataset to regress a difficult output to a realistic error for proof of concept for a larger project, incorporating a more significant dataset size and variety of both recipe and machine. Ideally, data science processes such as this could be accurate and robust across a wide variety of plasma etch silicon manufacturing processes and conditions to the degree that much of the process of recipe adjustment can be automated using ML pipelines. This project has accomplished its goals by resulting in a viable solution to the presented problem that tests the limits of practical machine learning.

\section{Acknowledgements}

We thank Lucas Frey, the engineer tasked with measuring the images that determined the output values for the model, which was crucial to the implementation of the model we settled on. Many others at Lam Research Corporation were instrumental to the success of this project and we extend our greatest appreciation to all those involved.

We thank Lam Research Corporation for providing both the data and the grant funding for this project and the BINDS Laboratory in the College of Information and Computer Sciences at the University of Massachusetts, Amherst for facilitating the project.

\newpage
\bibliographystyle{IEEEtran}
\bibliography{ms.bib}
\newpage
\begin{appendices}
\section{Hyperparameters Appendix}\label{appendix:hypers}

\begin{table}[h!]
    \centering
    \begin{tabular}{|l|l|}\hline
    \textbf{Attribute} & \textbf{Value} \\\hline
    Layers & 0, 1, 2, 3\\\hline
    Hidden units & 32, 125, 252, 350, 1e3, 2e3, 3.5e3, 4032, 5e3\\\hline
    Activation function & Hyperbolic Tangent, ReLU\\\hline
    L2 Regularization Constant & 0, 1, 5, 50, 100\\\hline
    Learning Rate & 1e-2, 1e-3, 1e-4, 1e-5, 1e-6, 1e-7\\\hline
    Learning Decay & 0, 1e-10, 1e-9, 1e-8, 1e-7, 1e-6, 1e-5\\\hline
    Learning algorithm & Stochastic Gradient Descent, Adam \\\hline
    Epochs & 1, 10, 25, 50, 75, 100, 150, 500, 1e3, 1e4 \\\hline
    ES Minimum Delta & 0, 1e-100, 1e-20, 1e-10\\\hline
    ES Patience & 0, 1, 5, 10, 25, 50\\\hline
    \end{tabular}
    \caption{Enumeration of all hyperparameters compared to determine the settings for the model that generalized best as defined by the cross-validation procedure.}
    \label{tab:hypers}
\end{table}

\newpage

\section{Cross-Validation Errors Appendix}\label{appendix:errors}

\subsection{Remaining Mask}

\begin{table*}[h!]
    \centering
\begin{tabular}{|c|c|c|c|}\hline
Fold&Training Errors&Validation Errors&Measurement Errors (nm)\\\hline
0&-1.61190&-10.71951&3.27001 \\\hline
1&-2.17978&-3.46008&1.81944 \\\hline
2&-2.44279&-0.29940&0.47359 \\\hline
3&-2.20296&-3.69779&1.91397 \\\hline
4&-2.30286&-2.12411&1.37505 \\\hline
5&-2.20845&-3.37437&1.82897 \\\hline
6&-2.31366&-1.84986&1.30542 \\\hline
7&-2.26967&-2.35925&1.49089 \\\hline
8&-2.44939&-0.15040&0.33892 \\\hline
9&-2.41733&-0.58120&0.46082 \\\hline
10&-2.39607&-1.06960&0.89664 \\\hline
11&-2.35517&-1.33912&1.13999 \\\hline
Mean & -2.26250 & -2.58539 & 1.35948\\\hline
\end{tabular}
    \caption{The errors catalogued by the cross-validation procedure for the output variable of Remaining Mask. We notice a fair amount of variety in the measurement error, indicating that certain wafers are more extreme examples with respect to the dataset than others.}
    \label{tab:mask_errors}
\end{table*}
\subsection{Recess}

\begin{table*}[h!]
    \centering
\begin{tabular}{|c|c|c|c|}\hline
Fold&Training Errors&Validation Errors&Measurement Errors (nm)\\\hline
0&-239.70212&-62.31758&7.57505 \\\hline
1&-164.60648&-26.26657&4.76816 \\\hline
2&-216.93628&-13.48704&2.97182 \\\hline
3&-200.22007&-28.50959&5.23397 \\\hline
4&-218.38041&-11.94878&2.78657 \\\hline
5&-218.47845&-5.04322&1.89414 \\\hline
6&-196.88474&-19.16364&4.27496 \\\hline
7&-198.58153&-9.00566&2.70473 \\\hline
8&-163.59976&-6.82301&2.26224 \\\hline
9&-162.42805&-17.24353&3.91904 \\\hline
10&-162.77509&-12.34732&3.39943 \\\hline
11&-258.79387&-0.89413&0.90290 \\\hline
Mean & -200.11557 & -17.75417 & 3.55775\\\hline
\end{tabular}
    \caption{The errors catalogued by the cross-validation procedure for the output variable of Recess. We notice a fair amount of variety in the measurement error, indicating that certain wafers are more extreme examples with respect to the dataset than others.}
    \label{tab:recess_errors}
\end{table*}

\newpage

\section{Learning Curves Appendix}\label{appendix:curves}

\subsection{Remaining Mask}\label{curve:mask}

\begin{table*}[h!]
    \centering

\begin{tabular}{ccc}\hline\\[-0.25cm]
\includegraphics[width=0.3\textwidth]{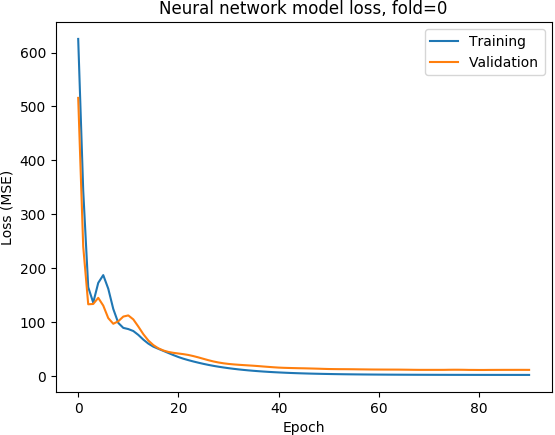}&
\includegraphics[width=0.3\textwidth]{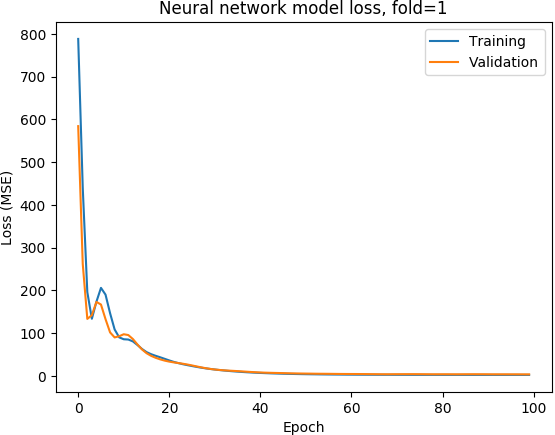}&
\includegraphics[width=0.3\textwidth]{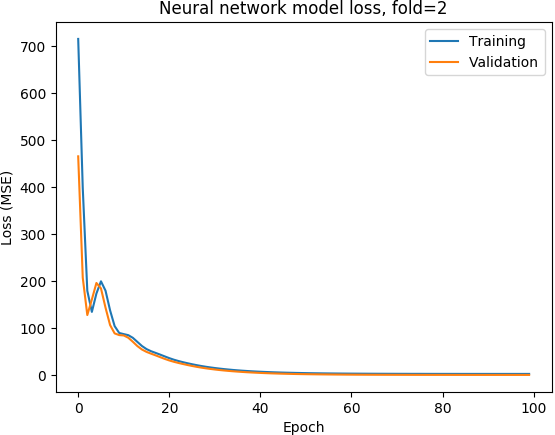}\\\hline\\[-0.25cm]
\includegraphics[width=0.3\textwidth]{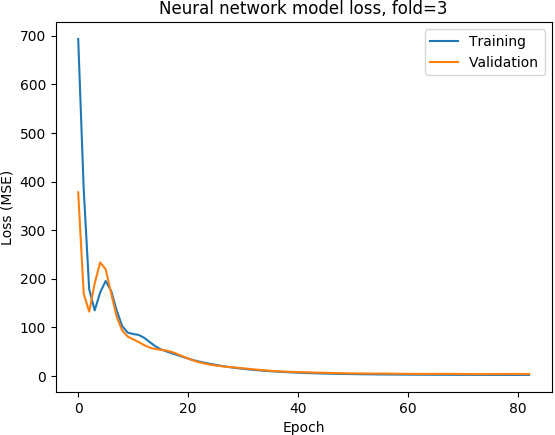}&
\includegraphics[width=0.3\textwidth]{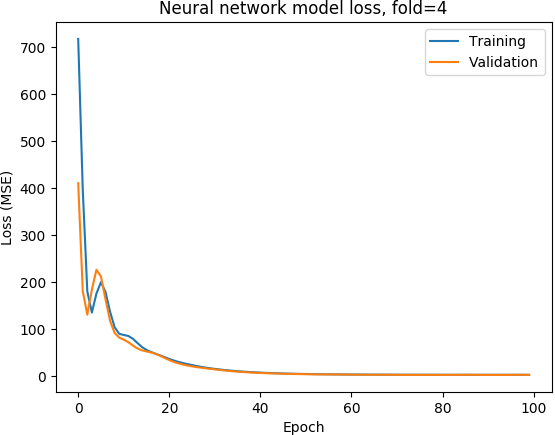}&
\includegraphics[width=0.3\textwidth]{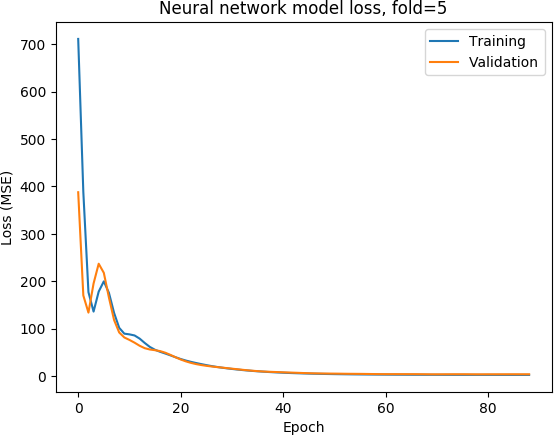}\\\hline\\[-0.25cm]
\includegraphics[width=0.3\textwidth]{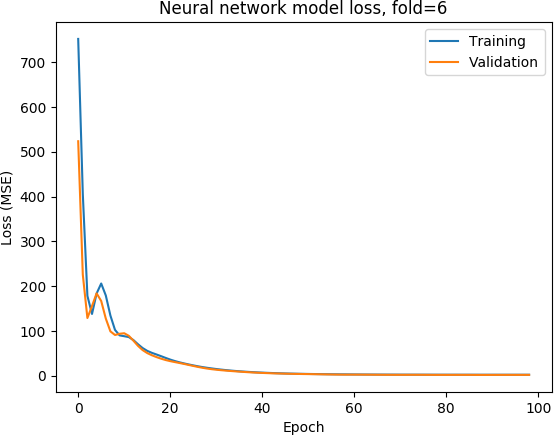}&
\includegraphics[width=0.3\textwidth]{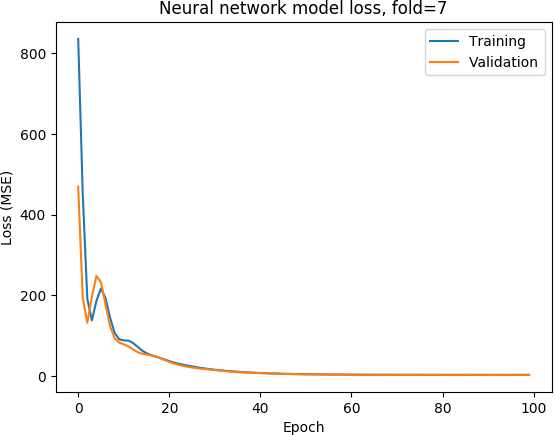}&
\includegraphics[width=0.3\textwidth]{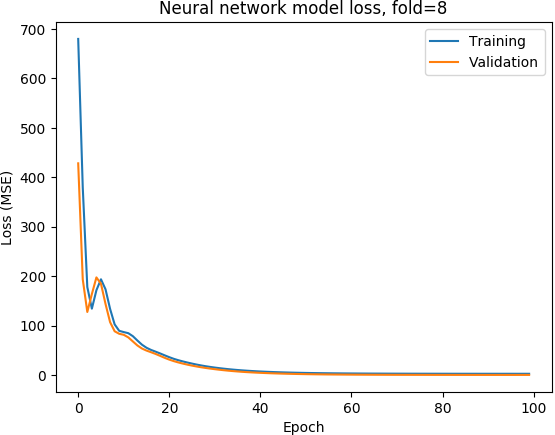}\\\hline\\[-0.25cm]
\includegraphics[width=0.3\textwidth]{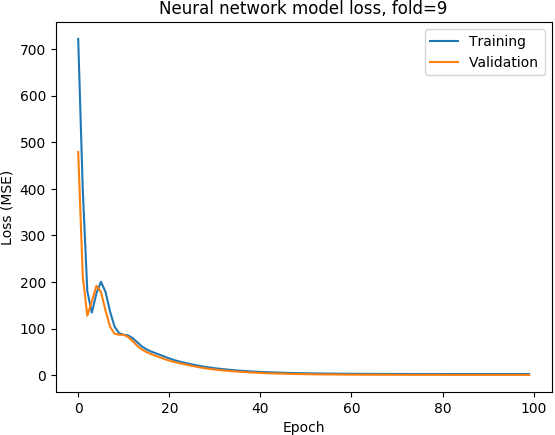}&
\includegraphics[width=0.3\textwidth]{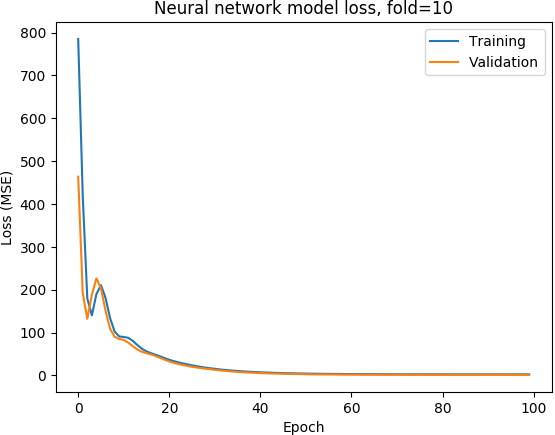}&
\includegraphics[width=0.3\textwidth]{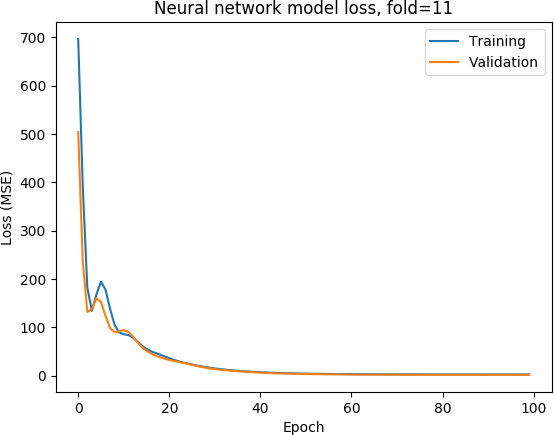}\\\hline\\[-0.25cm]
\end{tabular}
    
    \caption{Loss curves for each fold of training and validation mean squared error during the cross-validation regressing the Remaining Mask output variable. The orange represents validation error and the blue represents training error. As we can see above, the validation curve and the training curve both asymptote toward the Bayes error rate, and early stopping allows us to halt model training before the validation error rises again. In this way, we avoid losing expected generalization capability by selecting a model that is too complex.}
    \label{tab:mask_curves}
\end{table*}

\newpage

\subsection{Recess}\label{curve:recess}

\begin{table*}[h!]
    \centering

    \begin{tabular}{ccc}\hline\\[-0.25cm]
\includegraphics[width=0.3\textwidth]{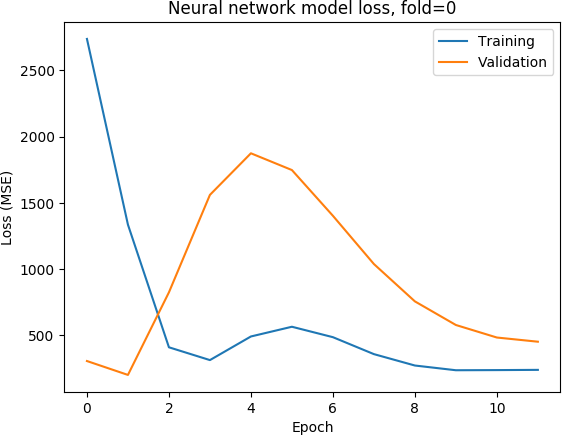}&
\includegraphics[width=0.3\textwidth]{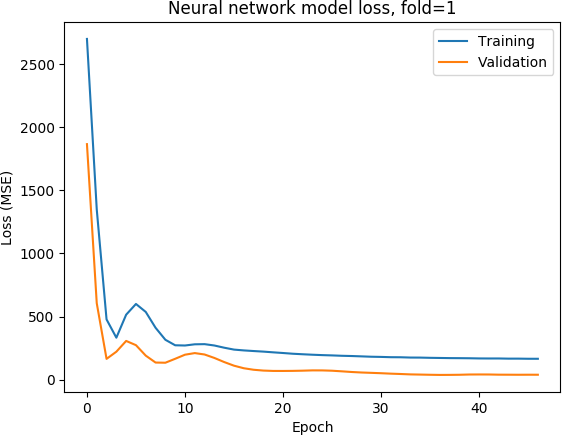}&
\includegraphics[width=0.3\textwidth]{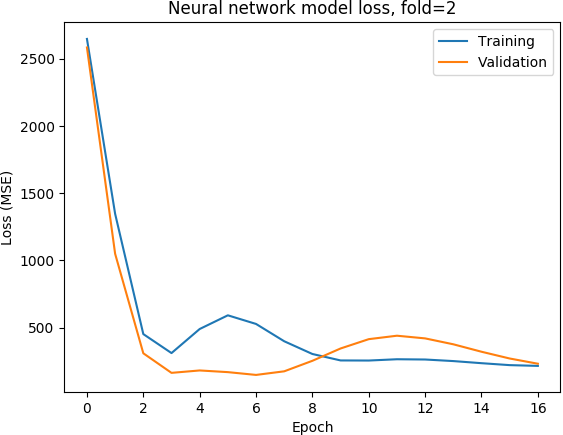}\\\hline\\[-0.25cm]
\includegraphics[width=0.3\textwidth]{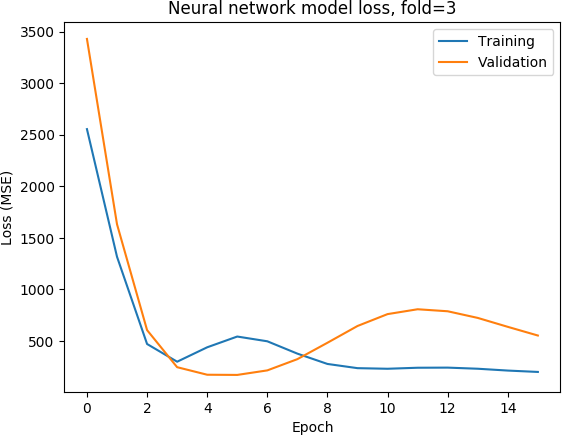}&
\includegraphics[width=0.3\textwidth]{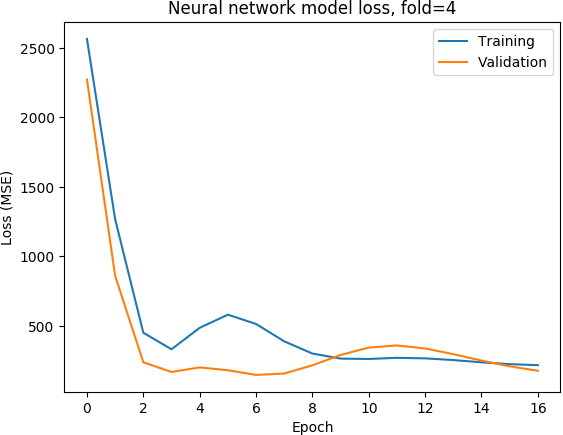}&
\includegraphics[width=0.3\textwidth]{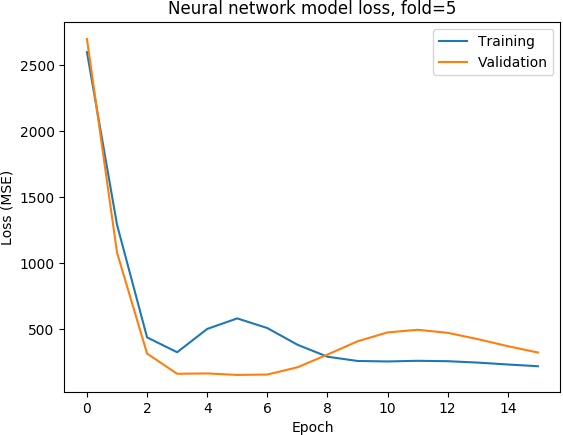}\\\hline\\[-0.25cm]
\includegraphics[width=0.3\textwidth]{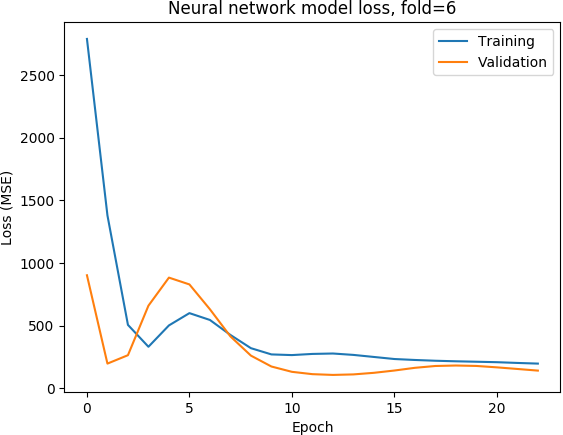}&
\includegraphics[width=0.3\textwidth]{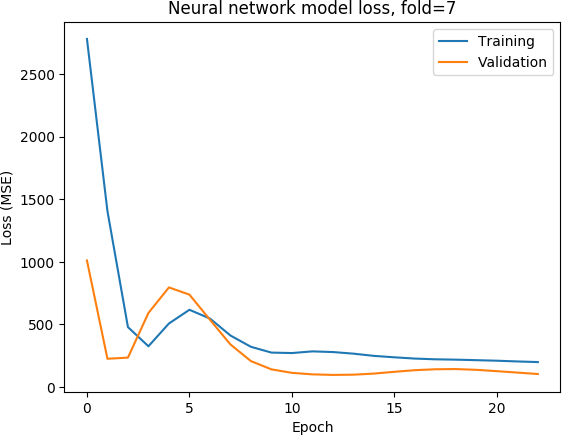}&
\includegraphics[width=0.3\textwidth]{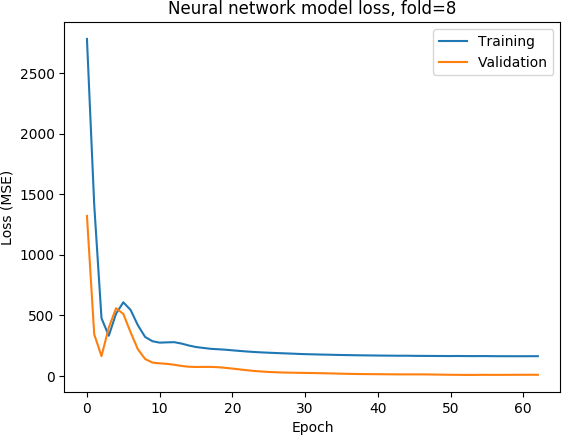}\\\hline\\[-0.25cm]
\includegraphics[width=0.3\textwidth]{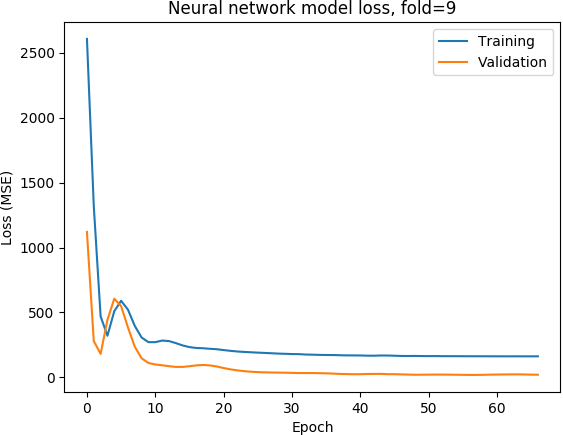}&
\includegraphics[width=0.3\textwidth]{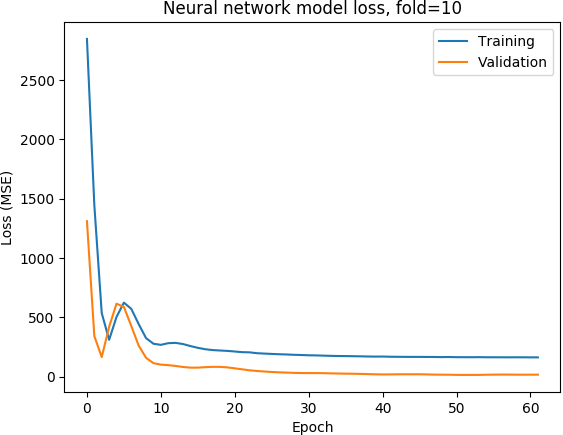}&
\includegraphics[width=0.3\textwidth]{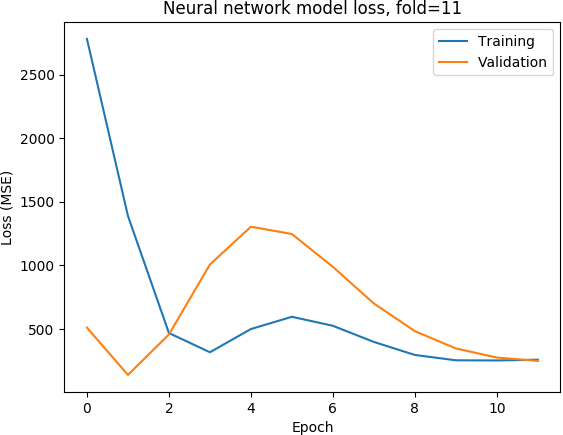}\\\hline\\[-0.25cm]
\end{tabular}

    \caption{Loss curves for each fold of training and validation mean squared error during the cross-validation regressing the Recess output variable. Notably, the hyperparameters selected are more well-suited for learning the Remaining Mask output variable, likely because of the limited scale of that variable. Future work includes, determining separate hyperparameters for each desired output variable, including the other two available not detailed in this report.}
    \label{tab:recess_curves}
\end{table*}

\end{appendices}

\end{document}